\documentclass[lettersize,journal]{IEEEtran}
\usepackage{amsmath,amsfonts}
\usepackage{algorithmic}
\usepackage{algorithm}
\usepackage{array}
\usepackage[caption=false,font=normalsize,labelfont=sf,textfont=sf]{subfig}
\usepackage{textcomp}
\usepackage{stfloats}
\usepackage{url}
\usepackage{verbatim}
\usepackage{graphicx}
\usepackage{cite}
\begin{document}

\title{V-LinkNet: Learning Contextual Inpainting Across Latent Space of Generative Adversarial Network}

\author{Jireh Jam, Connah Kendrick, Vincent Drouard, Kevin Walker, and Moi Hoon~Yap,~\IEEEmembership{Senior Member,~IEEE}
	% <-this % stops a space
	\thanks{Jireh Jam, Connah Kendrick and Moi Hoon Yap are with the Department of Computing and Mathematics, Manchester Metropolitan University, M1 5GD, Manchester, UK. E-mail: M.Yap@mmu.ac.uk.}
	\thanks{Vincent Drouard and Kevin Walker are with Image Metrics Ltd, Manchester, UK.}}% <-this % stops a space

\markboth{Journal (preprint) 2022}%
{Jam \MakeLowercase{\textit{et al.}}: V-LinkNet}


\maketitle

\begin{abstract}
%Deep learning methods outperform traditional methods in image inpainting. In order to generate contextual textures, researchers are still working to improve on existing methods and propose models that can extract, propagate, and reconstruct features similar to ground-truth regions. Furthermore, the lack of a high-quality feature transfer mechanism in deeper layers contributes to persistent aberrations on generated inpainted regions. To address these limitations, we propose the V-LinkNet cross-space learning strategy network. To improve learning on contextualised features, we design a loss model that employs both encoders. In addition, we propose a recursive residual transition layer (RSTL). The RSTL extracts high-level semantic information and propagates it down layers. Finally, we compare inpainting performance on the same face with different masks and on different faces with the same masks. To improve image inpainting reproducibility, we propose a standard protocol to overcome biases with various masks and images. We investigate the V-LinkNet components using experimental methods. Our result surpasses the state of the art when evaluated on the CelebA-HQ with the standard protocol. In addition, our model can generalise well when evaluated on Paris Street View, and Places2 datasets with the standard protocol.

Image inpainting is a key technique in image processing task to predict the missing regions and generate realistic images.
Given the advancement of existing generative inpainting models with feature extraction, propagation and reconstruction capabilities, there is lack of high-quality feature extraction and transfer mechanisms in deeper layers to tackle persistent aberrations on the generated inpainted regions.
% Despite the advancement of generative adversarial network brings to this field, the lack of a high-quality feature transfer mechanism in deeper layers contributes to persistent aberrations on generated regions. 
%We address this issue by proposing V-LinkNet, a dual-encoder method that exploits semantic coherency across textural features, with a new learning strategy for both encoders to communicate with each other. We design a loss function that employs both encoders to improve learning on contextualised features. We propose a recursive residual transition layer (RSTL) to extracts high-level semantic information. RSTL fuses the features from both encoders to exploit feature information at different scales.The V-LinkNet learns through latent space loss and adversarial loss to reconstruct images with similar pixel values of the target image. 
Our method, V-LinkNet, develops high-level feature transference to deep level textural context of inpainted regions our work, proposes a novel technique of combining encoders learning through a recursive residual transition layer (RSTL). The RSTL layer easily adapts dual encoders by increasing the unique semantic information through direct communication. By collaborating the dual encoders structure with contextualised feature representation loss function, our system gains the ability to inpaint with high-level features.
To reduce biases from random mask-image pairing, we introduce a standard protocol with paired mask-image on the testing set of CelebA-HQ, Paris Street View and Places2 datasets. Our results show V-LinkNet performed better on CelebA-HQ and Paris Street View using this standard protocol. We will share the standard protocol and our codes with the research community upon acceptance of this paper.

%intro
%method
%conclusion

\end{abstract}


\section{Introduction}
Recent advances in deep learning approaches have begun to dominate the field of algorithmic research in image inpainting. The Generative Adversarial Network based (GAN-based) technique to generate realistic images is possibly the most promising field in image inpainting. Besides GAN-based inpainting techniques, traditional methods \cite{efros1999texture,bertalmio2000image,barnes2009patchmatch} that employ propagation by pixel interpolation are still being researched. Image inpainting has demonstrated in many applications, which include image restoration \cite{wan2020bringing}, facial image editing \cite{jo2019sc}, facial wrinkle inpainting \cite{yap2021survey} and scene occlusion removal \cite{zhan2020self}. 

Image inpainting techniques has been classified into traditional methods and deep learning (learning-based) methods \cite{elharrouss2019image,jam2020comprehensive}. Traditional image inpainting methods propagate features from background or boundary regions to fill-in missing contents of damaged (foreground) or neighbouring regions. Depending on the contents of propagation, Jam et al. \cite{jam2020comprehensive} classified these methods into three categories, i.e., diffusion-based approaches, exemplar-based methods and hybrid methods.

The learning-based methods, popular known as deep generative neural networks, have become the state of the art, based on their ability to learn distribution with regards to context. These approaches \cite{pathak2016context,iizuka2017globally,liu2018image,yu2018generative,li2019progressive,liu2019coherent,zheng2019pluralistic,zhao2020uctgan,jam2020symmetric,jam2020r,jam2021foreground} use convolutional neural network (CNN) within an encoder-decoder within a GAN-based network to generate realistic images. These algorithms with a large amount of parameters and alternative layer configurations learn to manage feature extraction, propagation, and regularisation. However, the failures in generating contextualised features has probed the increased design of models, specifically targeting feature extraction and propagation capabilities in image inpainting \cite{wang2018image,xiao2019cisi} and other research domains \cite{chen2018double,chakravarty2019gen,chen2019image}. 

However, we also notice a gap in the research focus, the performance is always judged over existing models by a series of image quality measures, which could not as significant as projected, due to vulnerabilities in the qualitative analyses.
It is uncertain if these GAN-based inpainting algorithms can generalise well, if the same facial image is evaluated with different masks or if the same mask is applied to other facial images with varying contextual information can generate images that perform similarly.
The general question is whether there has been genuine progress in this field of research using GAN-based methodologies. This could be attributed to a variety of factors:
\begin{itemize}
	\item  Are the models reported using a standardised testing approach?
	\item Are these models being tested against a baseline model with predefined parameters?
	%   \item Are these models' baselines weak, or are newer models' baselines weak?
	\item Are papers explicit enough to improve reproducibility when comparing results? 
	\item Is the baseline utilised for comparison in the same domain?
\end{itemize}

\begin{figure*}[!ht]
	\centering
	\includegraphics[width=1.\linewidth]{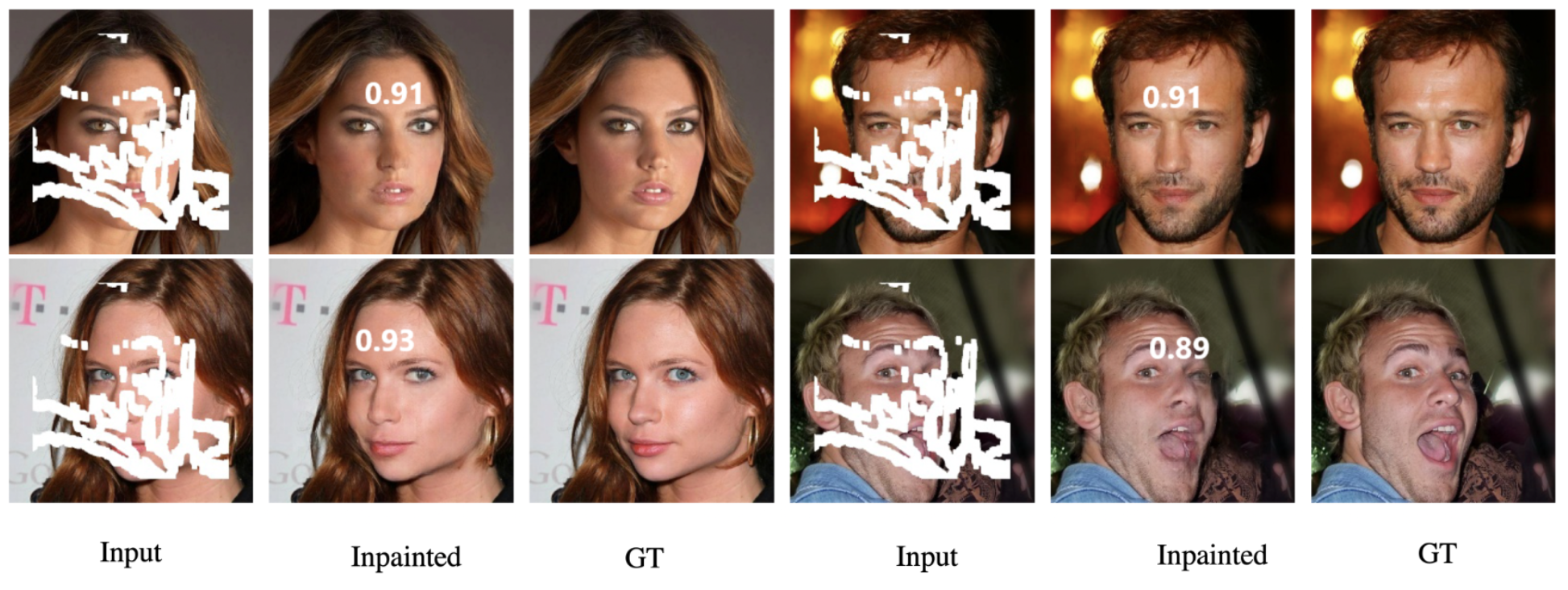} 
	\caption{\label{fig:v-gan_eval_faces} Illustration of inpainted results with one mask on different faces. Note the changes in SSIM values for different faces with the same mask.}
\end{figure*}

\begin{figure*}[!ht]
	\centering
	\includegraphics[width=1.0\linewidth]{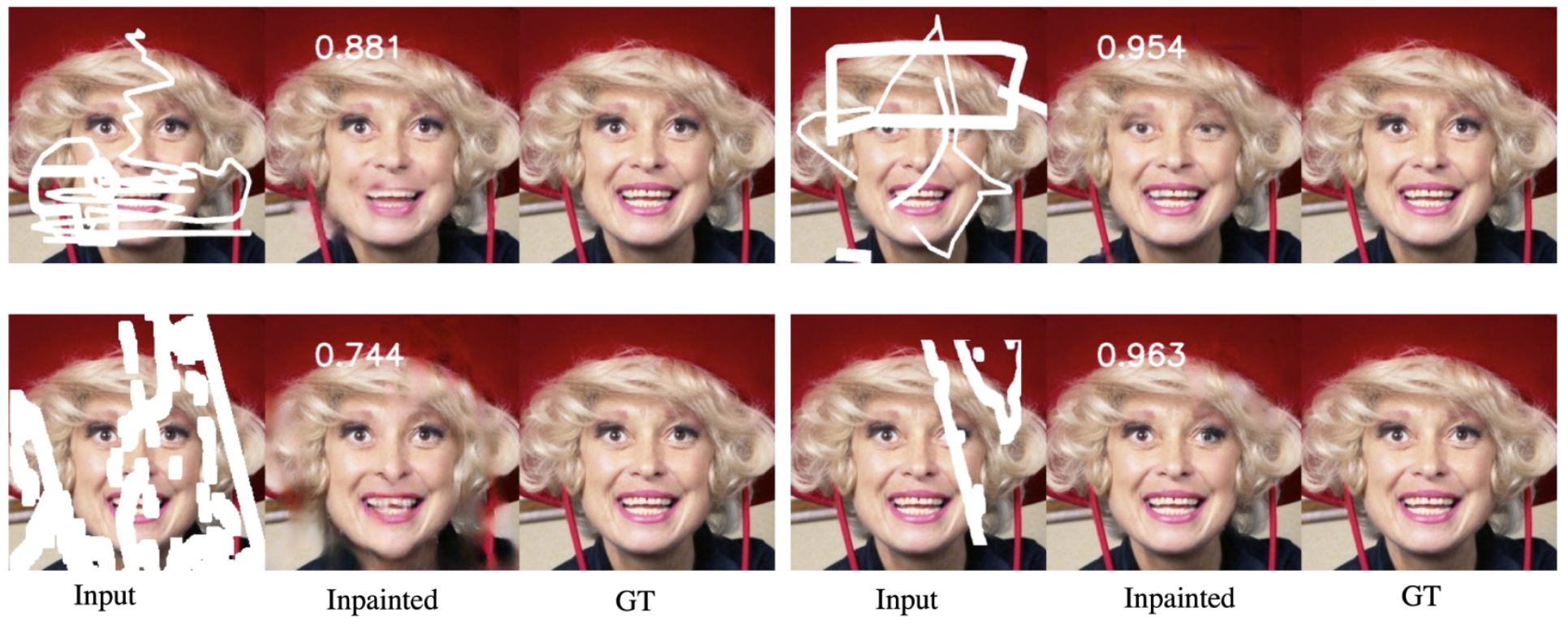} 
	\caption{\label{fig:ssimvalues} Inpainted images for the same image, different masks show different performance. See SSIM values on the inpainted images.}
	%The predictions of targeted regions show our model can image preserve realism.}
\end{figure*}

We observed that the datasets used for inpainting methods are often randomly split into training and testing sets. Also, the pairing of the masks are mostly random and lack a standard protocol. This is in vast contrast to other deep learning fields where training, validation and testing set are provided pre-split to ensure a robust and fair comparison. This is owing to the test set allowing an even split of cases under different conditions.
%Additionally, we identified biases in quantitative and qualitative evaluations with different masks and images with varying background and textural alterations. This is because image inpainting performance varies depending on image conditions. Lighting, backdrop, and facial features can all affect the prediction of the missing region. A face with a variety of illumination or posture change will under-perform.
Figure~\ref{fig:v-gan_eval_faces} illustrates the inpainting results of a mask on different faces using V-LinkNet. It is noted that the results are varied depending on the occluded regions. While most of the frontal and near to frontal faces achieved SSIM of 0.91 and above, the bottom right image achieved poor results with SSIM of 0.89, due to variation in lighting and facial expression. Figure \ref{fig:ssimvalues} shows another issue when we inpaint a face occluded by different masks. With different masks, the inpainted results have significant discrepancies; demonstrating vulnerabilities in the assessment pipeline where certain images and masks can be used to demonstrate high results. Therefore, algorithms performances are dependent on datasets and assessment approaches. While there is a rising trend in image inpainting research publications and codes sharing, we observe that there is no universal guideline for repeatable baseline result. This is due to the lack of standard protocol in this domain.

%-------------------------------------------------------------------------

In this paper, we address the research gap by proposing a cross-latent space reverse mapping GAN for image inpainting and a standard protocol to evaluate the performance of image inpainting. 
Our main contributions are:
\begin{itemize}
	\item We propose V-LinkNet, an end-to-end learning across latent space that uses feature information to encode fine details to complete the missing regions. We design a dual-encoder network approach and introduce a new learning strategy for both encoders to communicate with each other. This will improve networks internal collaboration allowing each encoder, to share features. Thus, distributing the task and focusing on unique high-level feature space representations.
	\item We design a RSTL to capture high-level features in a similar manner as maxpooling units within convolutions with feature preservation, and transfer technique employed as a ResNet-like unit within the block. This will allow the networks to extract and use high-level feature space representations in the inpainting task. Allow for increased detail in image reconstruction. 
	\item We introduce a standard protocol by pairing testing set images with masks, which will be made available for the research community.	This will facilitate a fair comparison of existing state-of-the-art image inpainting algorithms, and motivate reproducible research in the field of image inpainting.
\end{itemize}

We conduct an ablation study to validate the results of our proposed solution to image inpainting. We show that the results of the inpainting task can generate images with contextualised features.

\section{Related Work}
Image inpainting is an open and ongoing problem with extensive prior work in existence. This section summarises previous work with key focuses on GAN-based methods. For full review in image inpainting, refer to Jam et al. \cite{jam2020comprehensive}.

\subsection{Traditional Methods}
The main categories of traditional methods \cite{jam2020comprehensive} are summarised as:
\begin{itemize}
	\item Diffusion-based approach \cite{bertalmio2000image}. It transmits structural information from boundary areas into the interior, are among the categories of traditional inpainting techniques \cite{jam2020comprehensive}. Techniques in this category, on the other hand, produce blurry artefacts on large textured missing patches, which is undesirable.
	\item Exemplar-based methods \cite{efros1999texture,barnes2009patchmatch}. It uses similar patch searching techniques to fill in missing regions. Methods in this category attempt to address the limitation of diffusion methods on large textured regions, but still fail to match exact content and sometimes suffer with misalignment due to patch overlap, such as PatchMatch \cite{barnes2009patchmatch}. In addition, they are usually computationally expensive and time consuming with unrealistic results for large image-to-hole ratio with an arbitrary mask. 
	\item Hybrid methods. It uses both diffusion and exemplar-based methods to address misalignment by adding blurry effect in the boundary areas of the target region. Despite the success in producing textural features for a missing region, there are still issues with computation and aberrations persist. Nonetheless, the failure to capture high-level image features and the inability to generate complex and non-repetitive structures \cite{zeng2019learning,xie2019image} continue to be problematic.  
\end{itemize}

Reconstruction of complicated textural regions such as faces was also a challenge for traditional techniques. However, despite the reasonable results obtained with other natural scene images, the limited amount of high-level information accessible during computation has resulted in techniques in this category failing to produce high-quality features with believable semantic structures in natural scene images.
Traditional inpainted images frequently exhibit broken or unconnected edges along border regions, blurry artefacts, and overlapping patches along seam regions. Additionally, the inpainting process for techniques in this category is rather computationally intensive. 

\subsection{GAN-based Methods}
GAN-based methods make use of large-scale data to facilitate hallucination and the extraction of high-fidelity features  within CNN blocks for image inpainting. GAN models for image inpainting \cite{wang2018image,liu2020rethinking,yang2021mse} have used multi-columns to encode and propagated features directly to the decoder or use a self-supervised Siamese style inference approach \cite{xiao2019cisi}, where a style encoder is the supervisor of the generator, to improve feature extraction and learning. Other methods \cite{lu2018image,yu2019free,jam2020r} observed that failures in feature extraction and propagation could be due to the irregular holes. To address the limitation, Liu et al. \cite{liu2018image} proposed an independent mask updating with partial convolutions to specifically target missing regions. Yu et al. \cite{yu2019free} proposed to use gated convolutions to gear the model towards learning soft mask of the irregular hole regions. More recently, Jam et al. \cite{jam2020r} proposed a reverse mask mechanism to specifically target missing regions whilst preserving the visible ones using a spatial preserving operation. 

Alternative approaches are two-stage models \cite{song2018spg,song2018contextual}, where a coarse version is generated at stage 1, and then used as the input to a refinement network at stage 2. The issue with this approach is inadequate information during reconstruction, due to larger target pixel region, hence a poor input for decoration at stage 2. It is still a challenge to reconstruct high-dimensional distribution from natural scenes than from aligned faces with no visible aberrations. A couple of reasons could be failures in feature propagation techniques or lack of refinement mechanisms in deeper layers to capture high-resolution feature maps. To address the aforementioned limitations, Pathak et al. \cite{pathak2016context} proposed a channel-wise convolution layer for feature propagation but the drawback is high computation and inefficient transfer of feature maps.

\noindent
\textbf{Attention-based inpainting in GANs}
The attention mechanism method is a frequently utilised tool in computer vision problems because of its ability to focus on key features. Improved segmentation, re-identification, captioning, and tracking performance have all been demonstrated to be beneficial \cite{wang2020multistage}. Using a two-stage network with a contextual attention layer, Yu et al. \cite{yu2018generative} demonstrated that the attention layer assist the model in finding tiny texture details across patches within the masked regions in order to gather high-level features during inpainting. In other approach, Yu et al. \cite{yu2019free} presents an attention layer and a soft gating approach as gated convolutions to learn soft mask from data in order to increase the performance even more. Sigmoid activation is used by the gating mechanism to convey realistic qualities by scaling features between [0,1] in order to achieve this.

Attention mechanisms \cite{yan2018shift,yu2018generative,wang2019musical,zeng2019learning} have also been considered in deeper layers or as transition between the encoder and decoder. It is noted that a bottleneck \cite{zhou2019vision,yang2019lafin,li2019localization} or a feature transfer mechanism like attention layers within deep layers of convolutions is often required when inpainting high-resolution images. Due to high-resolution images, convolutional outputs required large amount of GPU memory, thus resulting to an increase training time \cite{pathak2016context,lin2017refinenet,zeng2021aggregated}. Additionally, when extracting features from high-resolution images, some features may be lost during the operation. However, more detailed information about low-level features, such as edges, is frequently captured within the first few layers of the convolution. As a result, failure to consider prior semantic distributions leads in unusual textures on the generated image. One limitation of attention mechanism is that it increases computational cost \cite{yu2018generative} and does not generalise well in feature propagation. Liu et al. \cite{liu2019coherent} considered pixel consistency and proposed a module that searches previous patches to extract relevant features.
Thus feature extraction and propagation are important factors to consider at the design stages of the network. Although these are actively being considered, there are still gaps for new approaches. One reason is the limitation in information dissemination of high-level features caused by the design of attention layers.
\\

\noindent
\textbf{Image gradients in GAN-based inpainting}
Image gradients are often used by different image processing techniques \cite{zhang2018deep,huang2019indoor}. This is due to the fact that the human vision is far more sensitive to gradients than it is to overall pixel intensity \cite{achanta2009saliency,wu2020histogram,sekehravani2020implementing}.
Because known and unknown areas are representations of the masked image in image inpainting, implementing a gradient method to identify occlusion boundaries might be beneficial.
Image gradients draw attention to directional changes in images and can be utilised in edge detection algorithms \cite{xie2015holistically}. Image gradients have been used in image inpainting \cite{bertalmio2000image,criminisi2004region,vo2018structural} by utilising edge information, which has been shown to be effective. It is possible to utilise diffusion-based inpainting to spread information around the borders of a painting from known to unknown regions by utilising fluid dynamics and partial differential equations. Because edges are continuous, information travelling through isophote (a line linking locations with the same pixel level intensity) matches gradient vectors at the border between the missing pixels and the known pixels. The usage of edge information, on the other hand, varies depending on the hyperparameter and the edge detector used. 

When it comes to image inpainting, the Sobel operator is not a new technique \cite{sadowski2018image}. Because images include noise, which may induce a rapid change in pixel values \cite{zhang2018deep}, the Sobel algorithm \cite{kanopoulos1988design} is capable of extracting occlusion boundaries. When employing the Sobel operator for edge detection, noise may be subdued without losing edges, edges can be improved by applying a high pass filter, and spurious edges, which are caused by noise, can be eliminated (edge localisation).
Sadowski et al. \cite{sadowski2018image} employed the Sobel operation to collect gradient information from generated and ground-truth images in order to construct a loss function. With the help of edge information, Zhang et al. \cite{zhang2019gain} was able to obtain gradient features that were later fused with image features to obtain the final image. They used a masked gradient map and mask to enable the network to obtain gradient features, which were later fused with image features to obtain a final image.
% We target the image gradients of feature maps in order to establish the direction of filling priority, and we utilise this information to create our loss model.
Several techniques have been developed in an attempt to apply structural restrictions to the inpainting tasks, such as two-stage networks, instance images, and matching completion images.
Reconstructed images, not with standing their successes, fall short of capturing high-level feature information of the target regions. This problem continues to be difficult, and there is still much opportunity for progress in this field of research.

\section{Method}
\subsection{Problem Formulation}
\begin{figure*}
	\centering
	\includegraphics[width=1.\linewidth]{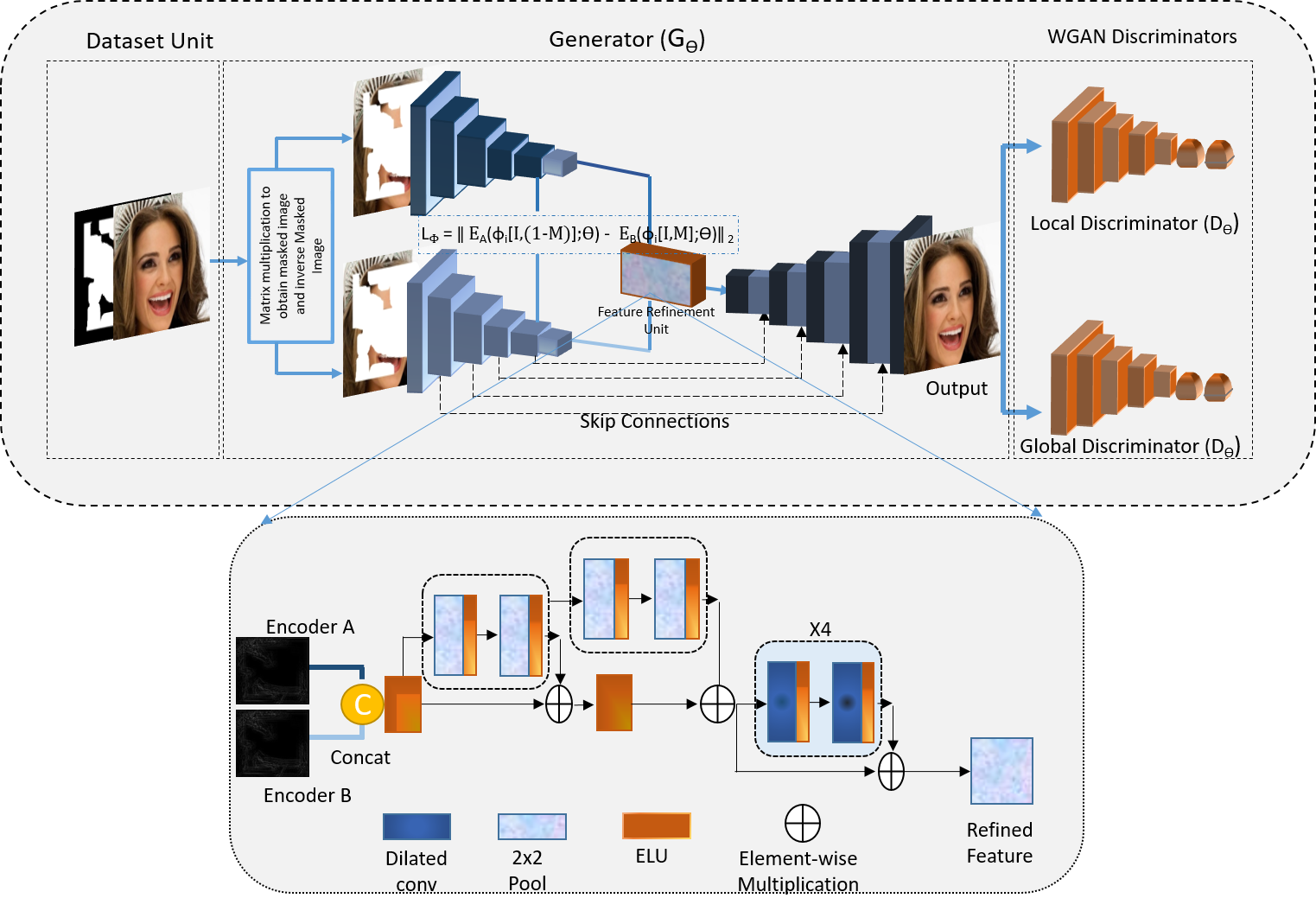} 
	\caption{
		Overview of our proposed architecture during training. The proposed RSTL passes refined features to learnable upsample layers within the decoder ($D_{\theta_{E}}(\cdot)$). The bottom diagram further illustrate the connected residual pooling. We utilize maxpooling with pool-size of $2 \times 2$ and ELU activation function as gating. The connected residual network uses dilated convolutions for refinement.}
	\label{FIG:VLinkStruct}
\end{figure*}

Previous works \cite{pathak2016context,iizuka2017globally,liu2018image,yan2018shift} in computer vision have shown that inpainting is a learning problem that can be solved by encoding high-level features.
The reconstructed output is geared towards having a close similarity to the input. We consider the inpainting task to have an input source $M_{I} = I \odot M$ and a target image $I$, where $M$ is a binary mask and $\odot$ is element-wise multiplication. 
The proposed V-LinkNet is a neural network generator with dual encoders of differing weights, a recursive
residual transition layer (RSTL), and a decoder. It utilises a global and local Wasserstein discriminator to build a generative model. We have included descriptions of the proposed technique as shown on (Figure~\ref{FIG:VLinkStruct}) with the RSTL for more clarity.
%to visualise the process during training. 
Additionally, we explain the training procedure that occurs to optimise the training loss functions.

\subsection{V-LinkNet Architecture}
Our proposed V-LinkNet is a generative model consisting of a generator, a global discriminator and a local discriminator as shown in Figure~\ref{FIG:VLinkStruct}. The discriminators are included for adversarial training. Only the generator network is used during the testing phase.

The generator $G_{\theta}$ has dual encoder branches ($E_{\theta_{A}}(\cdot)$ and $E_{\theta_{B}}(\cdot)$) and a decoder ($D_{\theta_{E}}(\cdot)$). Within $G_{\theta}$, encoder branch $E_{\theta_{A}}(\cdot)$ focuses on the capturing contextual information covered by the masked (unknown) regions. To ensure the reconstructed image is visually coherent with the structure and context of the ground-truth, we design $E_{\theta_{B}}(\cdot)$ to capture encoding with main focus on perceptual and structural information.

Both encoders $E_{\theta_{A}}(\cdot)$ and $E_{\theta_{B}}(\cdot)$ have eight convolution blocks, each with variations in spatial resolution and receptive fields at dilation rates of 2, 4, 8, 16. $E_{\theta_{A}}(\cdot)$ has dropout layers with value $0.2$ after each convolution block to reinforce learning.
Blocks one to five, have batch normalization and Exponential Linear Unit (ELU) activation followed by maxpooling layer, while blocks six to eight has ELU and dropout. 
Within the decoder $D_{\theta_{E}}(\cdot)$, are learnable upsampling layers using bilinear interpolation each with a convolution block that includes batch normalization and ELU activation layers. 
The final convolution block of the encoder-decoder (generator) has a Tanh activation layer with no batch normalization layer, which is deliberate so as to accelerate training and stabilize learning.
The output of the final layer $I_{pred} = D_{\theta_{E}}(g_{\theta})$ and the generator output as $G_{\theta}(I_{pred})$. The final output a generated image based on nonlinear weighted upsampling in latent space.

We train the V-LinkNet on the training set of $I$ and use high-level features within both encoders to minimise the error. Midway between the paired encoders, we modify the convolution block by increasing the dilation rate to 8 and 16. Both encoders  $E_{\theta_{A}}(\cdot)$ and $E_{\theta_{B}}(\cdot)$ learn high-level features to obtain output features $E_{\theta_{A}}(\phi)$ and $E_{\theta_{B}}(\phi)$, which are passed into a RSTL. 
The RSTL is designed to fuse the features from both encoders to exploit feature information at different scales.
The V-LinkNet learns through latent space loss and adversarial loss to reconstruct images with similar pixel values of the target image.
V-LinkNet consists of a training and inference (testing) phase. We use the traditional WGAN training, where the training sample are masks and ground-truth images.
During training, the network learns with the main objective being to generate an image given the mask and the ground-truth.
To minimise the error through back propagation, we design a new loss function that evaluates the training set to minimise the error in order to find high-level matching features between the paired encoders. The details of our proposed loss function will be discussed in Section \ref{section:loss}.
We project the corrupted input onto the latent space of the generator through iterative backpropagation. Therefore, we reuse the weights of both encoders to compute an objective function that will specifically target valid regions. At each stage of the network training, the weights will assist with fast updating during learning to guide the model.

\subsection{Recursive Residual Transition Layer}
\label{sec:unit}
Residual learning has been well established in deep learning due to their ability to reduce training error in much deeper layers. 
A simple implementation of a residual block is a fast-forwarded activation layer within the neural network. 
By adding the activation layer of a previous layer, to a deeper layer within the network, a residual connection is achieved. 
In previous works \cite{pathak2016context, yang2017high, jam2020symmetric}, feature extraction and propagation often fail with large portions of the background due to low level capture and poor transition to the decoder.
%not fully captured during feature abstractions within the decoder. 
We consider maxpooling, an operation that highlights the most present feature of an image patch and calculates its maximum value. 
Because features encode spatial representation of visible patterns, it is more informative to consider the maximum presence of different features extracted from the image. 
Hence the reason why maxpooling is considered instead of average pooling in this work. 

% \begin{figure}
% 	\centering
% 	\includegraphics[width=1.\linewidth]{fig5.png} 
% 	\caption{
% 		Illustration of connected residual pooling. We utilize max-pooling with pool-size of 
% 		$2 \times 2$ and ELU activation function as gating. The connected residual network uses dilated convolutions for refinement.}
% 	\label{FIG:RSTL}
% \end{figure}

We design the RSTL, as illustrated in Figure \ref{FIG:VLinkStruct}, with the aim to capture high-level semantic information. There are two units in the RSTL: a maxpooling residual connected unit and a convolution residual connected unit. The RSTL is formed by residually connecting both units. The idea is to efficiently pool multiple window sizes, combine them using learnable weights and fast-forward to deeper layers to reduce the error during training. As a result, training gradients are obtained from the next connected layer within each layer, and these gradients are used to update the parameters in the current layer. This, in turn, influences the weights of the filters, causing the activation maps to increase or decrease, lowering the loss.
We combine the residual connection of the activation maps with the output of the final pooling layer and the input of the residual layer to obtain the unit output feature map. To feed the RSTL, we concatenate the output feature maps of both encoders.
The RSTL extracts high-level semantic information from the concatenated input to generate a feature map, which is then passed on to the decoder.

Our proposed connected pooling operation combined with residual connections reduces the error near the boundary regions of the hole regions, as it captures fine contextual information. 
This unit is designed to predict and delineate any mask residues as it highlights high-level semantic information by recursively performing pooling and convolution operations.
The concatenated features when passed via pooling unit suppresses noise to project informative pixels.
This is different from using the channel-wise attention which squeezes the spatial dimension of the feature map.
The objective of this design is to extract meaningful pixels whilst suppressing uninformative ones before passing them to the decoder. 
First we concatenate features extracted from $g_{\theta_{A}}(\cdot)$ and $g_{\theta_{B}}(\cdot)$ and pass them through ELU activation and then perform pooling operation followed by a $3 \times 3$ convolution and ELU unit as a gating layer. 
For more refined details we use a dilation rate of 16 within the convolution layer. 
\begin{equation}
	\label{eq:attentionguidedfeaturemap}
	F_{\phi}(X_{i}) = \sigma ([M_{pool}]F_{3 \times 3})
\end{equation}
where $\sigma$ is the ELU activation function, $F_{3 \times 3}$ is the convolution layer. The final feature map is given by:
\begin{equation}
	\label{eq:attentionfeaturemap}
	X_{\phi} = F_{\phi}(X_{i}) \oplus X_{i}
\end{equation}
where $\oplus$ is element-wise addition and lastly a dilated convolution layer to refine and transfer the feature to the decoder.

\subsection{Loss Function}
\label{sec:ch7loss}
To optimise the RSTL and dual encoders, we introduce a novel loss function that uses features of both encoders to assist the model during learning.
To ensure high-level contextual features for missing regions, we introduce a loss between $E_{\theta_{A}}(\cdot)$ and $E_{\theta_{B}}(\cdot)$.  
During training, the loss between both encoders ensures ongoing communication, in order to improve the models learning on contextual information. By employing this technique, the model can enhance visual consistency with contextualised features. The loss model is designed based on the Mean Squared Error (MSE) specifically to penalize large errors and provide fast learning. 
%Also included are other losses used during training of the V-LinkNet model. 

\noindent
\textbf{Feature Losses}
More recent approaches \cite{yu2019free,lu2020semantic,yi2020contextual,zeng2021aggregated} use pre-trained VGG16 or VGG19 \cite{simonyan2014very} to evaluate or enhance the perceptual quality of image inpainting results. 
These models \cite{wang2018image,yu2019free,zhao2020uctgan,yi2020contextual} have perceptual and style losses and these losses are still undisputed when it comes to evaluating or improving the overall performance of the generator.
Inspired by perceptual losses in feature space, we propose a novel feature loss in latent space. 
Features are low-dimensional latent state representations captured in latent space.
By reusing deeper-level features from both encoders in latent space, we design an objective learning loss model to capture rich features of the reversed regions covered by the mask. 
In addition, it is desirable for the inpainted regions to be as close to the counterpart regions of the ground-truth. 
Thus a head-start with faster update of parameters and weights to the generator is important in this task.
The reasons for this is that both encoders will learn from each other. 
Another reason is that loss functions can become difficult in latent representations, thus reusing latent representations to compute error within layers give the network easy access to compute the gradients for better head-start. 
This, enables understanding of contextual features during learning from the reversed input regions for a reasonable prediction. Hence continuous training can potentially capture subtle or more refined features in space. However, using all layers to find a better gradient computation increases computational complexity hence why only two identical deep layers of both encoders are used for this experiment.
% Thus, enabling understanding of contextual features during learning from the reversed input regions for a reasonable prediction.

% Moreover, there is no guarantee that continuous training will accurately capture more refined features in space for this specific task. 
% Thus using all layers to find a better gradient computation increases computational complexity hence why only one deeper identical layers of both encoders are used for this experiment.

\noindent
\textbf{Latent space feature-aware gradient loss}
Utilising image gradients is a very common practice of various image processing algorithms \cite{zhang2018deep,huang2019indoor}. The Sobel operator \cite{kanopoulos1988design} (filter) is a gradient operator that measures gradients on 2D images by capturing focused information. It works by directing attention to areas of high spatial frequency that correspond to the image's edges.
\begin{figure*}%[!h]   
	\centering
	\includegraphics[width=0.8\linewidth]{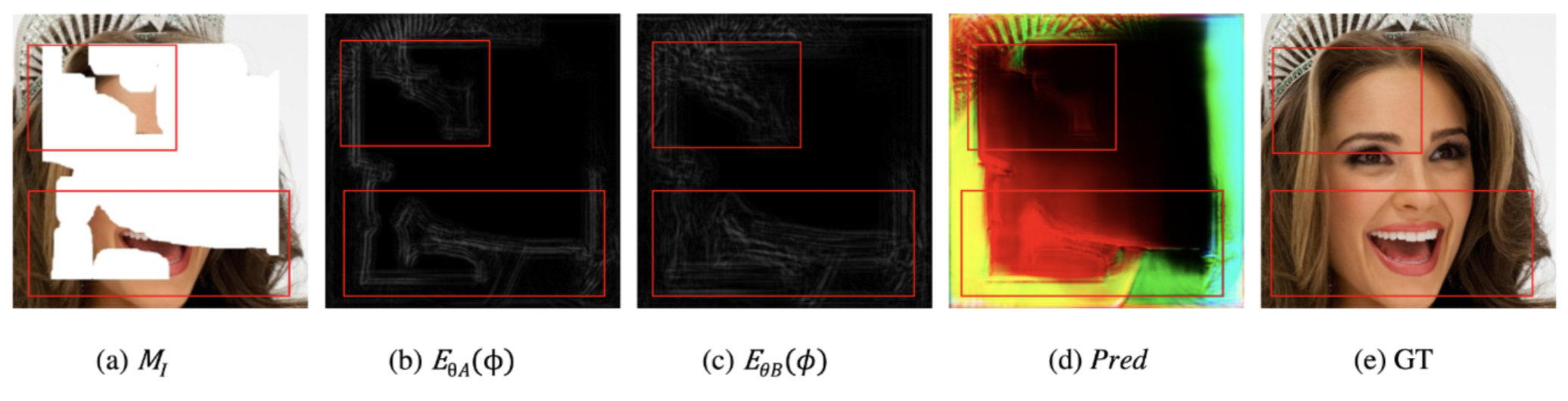} 
	\caption{Visual changes using Sobel operator \cite{kanopoulos1988design} on an image, where (a) is the masked input (b) $G_{xy} E_{\theta_{A}}(\phi)$ is the Sobel operator on the third convolution layer output of $E_{\theta_{A}}$. (c) $G_{xy} E_{\theta_{B}}(\phi)$ is the Sobel operator on the third convolution layer output of $E_{\theta_{B}}$. Notice the mask regions during convolution. The pixels with same intensity are black while neighbouring pixels that differ strongly are white. Compare output of the same layer of both encoders. Notice the difference in convolutional output. (d) First Prediction. Improves over time during training. (e) ground-truth image.}
	\label{fig:sobels}
\end{figure*}
In image inpainting, known and unknown regions are representations of the masked image, thus applying gradient algorithm to detect occlusion boundaries can prove useful. To determine the direction of filling priority, we target the image gradients of feature maps and use this information to construct a loss model.

We obtain feature gradients of the third convolutional layer of both encoders and compute the loss model. 
To re-enforce on outer edges and fidelity of the generated image, we utilize gradients of the generated and ground-truth images to assist in the final reconstruction. Note that the edge map based on the gradients are computed in x and y directions.

\noindent
\textbf{Generator Loss}
\label{section:loss}
The generator loss evaluates the missing pixel region and the perceptual quality of the image. 
To maximise contextual and feature-wise learning, we extract high-level features from deeper layers of $E_{\theta_{A}}(\cdot)$ and corresponding features of $E_{\theta_{B}}(\cdot)$. 
 
\begin{equation}
	L_{\phi} =  || (E_{\theta_{A}}\phi[M,I_{gt}] -E_{\theta_{B}}\phi[(1-M),I_{gt}])|| ^2_2 
	\label{eq:latentloss-function}
\end{equation}

\begin{equation}
	G_{x} = I_{gt} \ast X_{edge}(i,j), G_{y} = I_{gt} \ast Y_{edge}(i,j)
\end{equation}
where $G_{x}$ and $G_{y}$ are gradients computed by depth-wise convolution using the x and y components of the Sobel operator on an image $I_{gt}$. 
\begin{equation}
	\nabla I_{gt} = \sqrt{G_{x}^2 + G_{y}^2}
\end{equation}
\begin{equation}
	\label{eq:edgloss}
	L_{edge} =||\nabla I_{gt} -[G_{\theta}({\nabla I}_{pred})]||^2_2  
\end{equation}
\begin{equation}
	\label{eq:featureEdge}
	\nabla \phi =||\nabla E_{\theta_{A}}\phi[M,I_{gt}] -E_{\theta_{B}}\phi[(1-M),I_{gt}]||^2_2  
\end{equation}
\begin{equation}
	L_{edgeLoss}  = \lambda L_{\phi}+(1-\lambda) L_{edge}
\end{equation}
where $\lambda = 0.5$ as coefficient to obtain $L_{edgeLoss}$.
We use pixel space L1-norm, based on a range of pixel values with the input image and output image.
\begin{equation}
	\label{eq:ploss}
	L_{pix} =|| K \odot (I_{gt}(i,j) - I_{pred}(i,j))|| ^1_1 
\end{equation}
%-------------------------------------------------------------------------
\begin{equation}
	L_{vgg} =  ||\phi[I_{gt}] - \phi[G_{\theta}(I_{pred})]|| ^2_2 
	\label{eq:vggloss-function}
\end{equation}
%-------------------------------------------------------------------------
where, $K$ is a constant, $\odot$ is the element-wise multiplication, $I_{gt}$ is the ground-truth image and $I_{pred}$ is the predicted image. Further, we utilize the reversed mask loss ($L_{rm}$) from \cite{jam2020r} and compute a contextual loss. We want to keep the known pixel locations of the input image by penalizing the predictions thus creating similar pixels based on the reversed mask and masked regions.
%-------------------------------------------------------------------------
\begin{gather}
	I_{c} = M \odot I_{gt} + (1-M) \odot G(z) \\
	L_{rm} =  ||\phi(1-M \odot I_{gt}) - \phi({I}_{c})||^1_1
	\label{eq:comp}
\end{gather}
%-------------------------------------------------------------------------
\begin{equation}
	L_{c} =  ||\phi(1-M \odot I_{gt}) - \phi({I}_{pred} \odot M||^2_2
	\label{eq:plossn}
\end{equation}
The total loss ($L_{T}$) is a weighted sum of all the losses with highest weight applied to $L_{\phi}$. %-------------------------------------------------------------------------
\begin{equation}
	L_{T} =  \alpha_{1} L_{vgg} +  \alpha_{2} L_{rm} + \alpha_{3} L_{pix}
	\label{eq:total-function}
\end{equation}
where $\alpha_{1} =0.5,  \alpha_{2} =0.3,  \alpha_{3} =0.1$ are coefficients of the weights applied to the loss. 
% The state-of-the-art \cite{liu2018image,guo2019progressive,yu2019free} have used feature space. However, pixel space utilization is not new to inpainting. We realised that, using pixel space and edge-based loss reduces the checkerboard artefacts during upsampling. 

\noindent
\textbf{Discriminator Loss}
We utilise the Wasserstein distance loss similar to \cite{wang2018image,jam2020r} in both discriminators.
\begin{equation}
	{L}_{WGAN} = E_{I\sim P_{x}}[D_{\theta}(G_{\theta}(I_{gt}))]- E_{I_{pred}\sim P_{z}}[D_{\theta}(G_{\theta}(I_{pred}))]  
	\label{eq:W-distance-loss1-function}
\end{equation}
where real-data distribution is represented in the first term and generated-data distribution is the second term. The local discriminator $L_{D_{l}}$ uses the same loss term as the global discriminator $L_{D_{g}}$, but only provides loss gradients for missing regions during training.
The final objective loss for the discriminator is:
\begin{equation}
	L_{adv} = L_{D_{g}} + L_{D_{l}}
\end{equation}
% \begin{equation}
% {L}_{w} = E_{I\sim P_{x}}[D_{\theta}(G_{\theta}(I_{gt}))]- E_{I_{pred}\sim P_{z}}[D_{\theta}(G_{\theta}(I_{pred})\odot 1-M)]  
% \label{eq:W-distance-loss1-function}
% \end{equation}
Finally, we combine the objective loss function of the model defined in Equation~\ref{eq:loss}.
\begin{equation}
	L_{F} = L_{T} + L_{adv}   
	\label{eq:loss}
\end{equation}

\section{Experiments}
\label{sec:ch7experiments}
In this section, we describe the implementation of V-LinkNet, the datasets and introduce a standard protocol to benchmark the performance of image inpainting algorithms.

\subsection{Implementation }
\label{section:implemetation}
Implementation of the V-LinkNet is done using the Keras library with a Tensorflow backend, and the model is trained on a P6000 GPU computer. We resized our images to $256 \times 256 \times 3$  and align them with random masks during training and during testing aligned them with appropriate masks.
We pretrain the network using a novel loss function that backpropagates gradients using features from both encoders. We use the RSMProp optimizer, with a 0.0005 learning rate. We updated the generator and discriminator networks following pretraining and utilised the Adam optimizer \cite{kingma2014adam} with a learning rate of 1E-4 and a beta of 0.5. The network is trained with a batch size of 5 and for 100 epochs, which takes around three to five days depending on the amount of the training data. After obtaining a well-trained model, we use reverse mask loss and a decreased learning rate (1E-5) to fine-tune it while retaining the original network topology. The input is updated throughout completion using a contextual loss and a perceptual loss with coefficients of 0.4 and 0.6, respectively. Stochastic clipping is employed during back-propagation. We picked a modest value to ensure that contextual loss is prioritised during test-time optimization, and that the inpainted part of the generated image most closely resembles the input background context of the entire image. The generator and discriminator are fixed during back-propagation. 
% Note that due to the limited computational resources, we do not pretrain the network for $512 \times 512$ images. 
We evaluate the performance of V-LinkNet on CelebA-HQ, Paris Street View, and Places2 datasets, which are the most widely used datasets by the state of the art.

A fully trained model can predict missing pixels for image-to-mask ratios ranging from [0.1 to 0.8] during testing. The inference time is between 0.192 seconds to 63 seconds depending on the mask size. During inference, we employ the network design with batch normalisation layers disabled.

\subsection{Datasets}
\label{sec:std_dataset}
The images and the masks are two essential components to train and test the performance of inpainting methods. The following are the most commonly used datasets to evaluate the performance of image inpainting algorithms: 
\begin{itemize}
	\item \textbf{CelebA-HQ} \cite{karras2018progressive}: A dataset curated from CelebA \cite{liu2015deep} containing 30,000 high quality images of $1024 \times 1024$, $512 \times 512$, $256 \times 256$ and $128 \times 128$ resolutions. We followed the state-of-the-art procedures \cite{liu2018image} and split our dataset into 27,000 training set and 3,000 testing set.
	\item \textbf{Paris Street View} \cite{doersch2012makes}: This dataset contains 14,900 training images and a test set of 100 images collected from Paris street views. The main focus of the dataset is the buildings of the city and very important in geo-location task.
	\item \textbf{Places2} \cite{zhou2018places}: A dataset containing over 1 million images from 365 scenery from places. It is suitable for model learning and understanding of diverse complex natural scenery.
	The following scene categories were chosen: butte, canyon, field, synagogue, tundra, and valley (in that order) as proposed by Yan et al. \cite{yan2018shift}. In each category, there are 5,000 training images and 900 test images. Our model is trained on the training set and evaluated on the testing set.
\end{itemize}
Each training image for both Paris Street View and Places2 is resized to $256 \times 256$ pixels, which is then used as an input to our model.

\subsection{Standard Protocol Testing Dataset}
\begin{figure}[!ht]
	\centering
	\includegraphics[width=1.\linewidth]{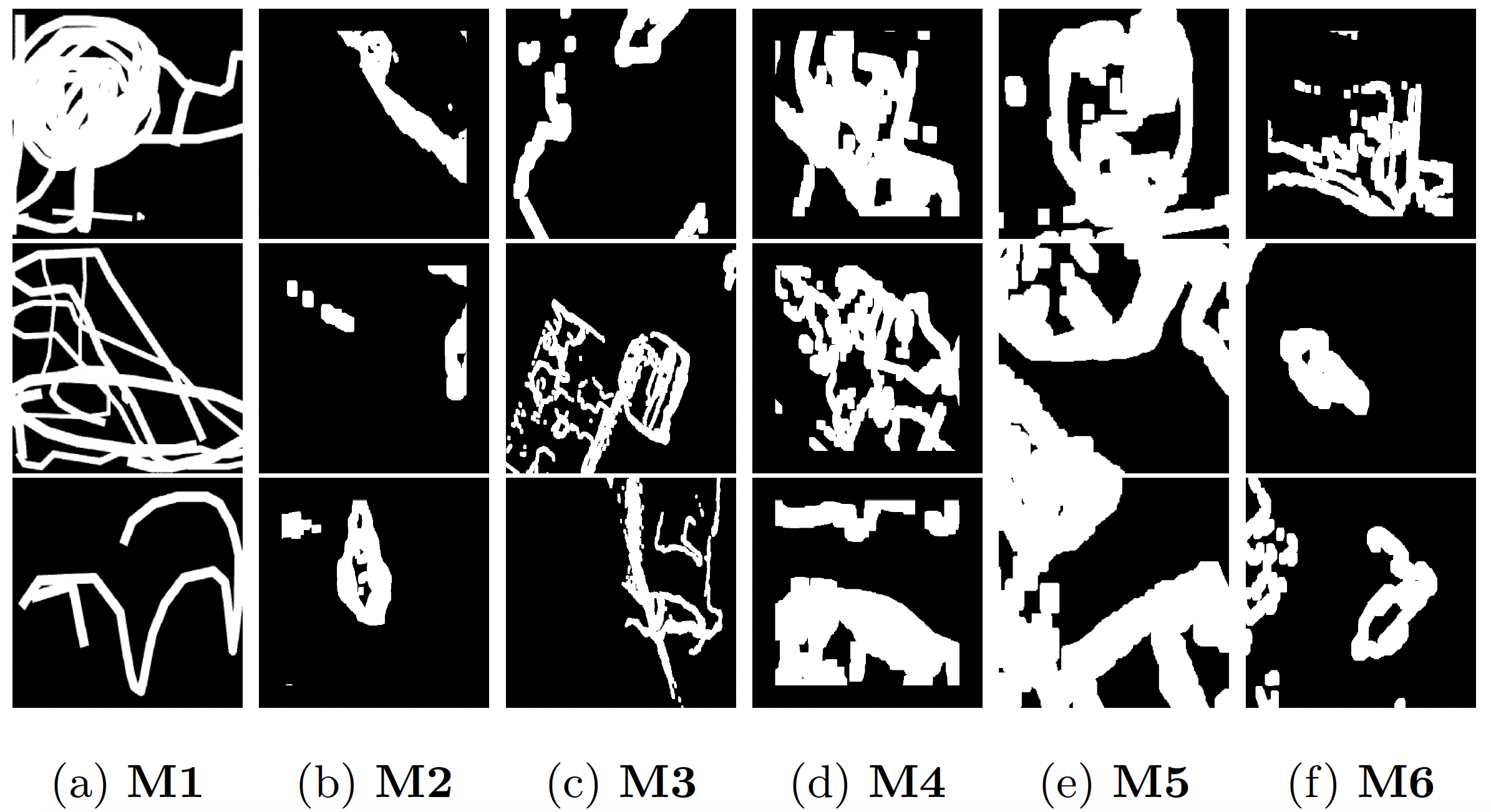} 
	\caption{Examples of the mask datasets. (a) MaskDataset1 [0.001,0.6] \textbf{M1} \cite{iskakov2018semi} (b) MaskDataset2 [0.001,0.1] \textbf{M2}  (c) MaskDataset3 [0.1,0.3] \textbf{M3}  (d) MaskDataset4 [0.3,0.4] \textbf{M4}  (e) MaskDataset5 [0.5,0.6] \textbf{M5}  (f) MaskDataset6 [0.1,0.4] \textbf{M6}. Note that \textbf{M2} to \textbf{M6} are from the Nvidia Mask dataset \cite{liu2018image}  }
	\label{fig:masksets}
\end{figure}

To encourage full reproducibility of our work, we introduce a standard protocol for image inpainting testing datasets. The facial test set is labeled according to CelebA-HQ \cite{karras2018progressive} dataset, which contains 3,000 high-resolution face images from CelebA \cite{liu2015deep}. The facial test set is split randomly according to \cite{liu2018image}. These images are paired with 3,000 masks images from Quick-Draw Mask \cite{iskakov2018semi} and 3,000 masks images from the Nvidia Mask \cite{liu2018image} test dataset, as illustrated in Figure~\ref{fig:masksets}. We will share the filenames of the paired image and mask test set in a comma-separated value (CSV) file. Note that our evaluation masks are set to 3,000 mask images and the images and masks used for training are not paired. The proposed standardised test dataset is curated with the facial image pose in mind. We evaluate the difficulty in inpainting task based on pose and variation in mask holes.

The Paris Street View \cite{doersch2012makes,pathak2016context} were standardised by Pathak et al. \cite{pathak2016context}, which is available upon request only from the authors. We adopted the Pathak et al.'s protocol for the Paris Street view dataset, which has 100 test images but used our own masks for testing. On the other hand, the Places2 \cite{zhou2018places} test dataset is extract from Places365-Standard. The categories used are butte, canyon, field, synagogue, tundra and valley. These are the same categories for training and testing. Each training set has 5,000 images, 900 test images and 100 validation images as per \cite{yan2018shift}. In total, there are a total of 5,400 test images. We paired each images with 5,400 masks and use it as a standard protocol for testing. For follow this due to longer training times and also based on the split by the state-of-the-art \cite{yan2018shift}.
%Figure~ \ref{fig:ynentresults} shows the evaluation of our paired standardised protocol specific to a facial image and the performance evaluation compared with the state of the art methods. 
For Places2 and Paris Street View datasets, we run evaluations based on the mask difficulty on their standard test set. 

\section{Results}
\label{sec:ygresults}
This section presents a quantitative and qualitative evaluation of the proposed V-LinkNet in comparison to state-of-the-art methods.

\subsection{Baseline model Comparison}
Without bias and dependent on codes availability, we used Pathak et al. \cite{pathak2016context} (\textbf{CE}), Liu et al. \cite{liu2018image} (\textbf{PC}), Yu et al. \cite{yu2019free} (\textbf{GC}), and Jam et al. (\textbf{RMNet}) \cite{jam2020r} as baseline models. The following summarise the baseline models used as the benchmarks for our standard protocol:
\begin{itemize}
	\item \textbf{Context encoder-decoder framework (CE)} \cite{pathak2016context} introduced the channel-wise fully connected layer to solve the convolutional layer limitation associated with failures in direct connection of all locations within a specific feature map. The channel-wise fully connected layer is designed to directly link all activation; thus enabling propagation of information within the activation of a feature map. 
	\item \textbf{Partial Convolution (PC)} \cite{liu2018image} proposed partial convolutions with mask updating to enforce learning in irregular hole regions during convolutions and ease feature transfer to subsequent layers, allowing convolution layers to target more of the missing regions as a result. 
	\item \textbf{Gated Convolution (GC)} The authors \cite{yu2019free} proposed a gating mechanism that learns soft masks within convolutions to make the transfer of features within convolutions more convenient. This method differs from PC \cite{liu2018image} in that the irregular mask is learnt rather than being updated in each step, whereas the former does not have this feature.
	\item \textbf{Reverse Masking Network (RMNet)} \cite{jam2020r} introduced reverse mask mechanism within the network. The reverse mask forces the convolutions to subtract visible regions through the reverse mask mechanism, thus ensuring the output prediction is on the missing regions only.
	% 	\item \textbf{V-LinkNet1}  (VN1) Introduced edge guidance in the form of Sobel loss during latent space learning. Expected results are not comparable to the state of the art baselines. We remove the pooling unit within recursive residual transition layer and train the model with this slight modification.
	%  	\item \textbf{V-LinkNet}
	% 	Pre-trained model with latent space loss and introduce recursive residual transition layer combined with perceptual losses and reverse mask loss. The recursive residual transition layer forces the model to focus on learning between disentangled pixels of valid and non-valid regions as it transfers highlighted high-quality features to the decoder.
\end{itemize}

\subsection{Qualitative Results}
\begin{figure*}[!ht]
	\centering
	\includegraphics[width=1.\linewidth]{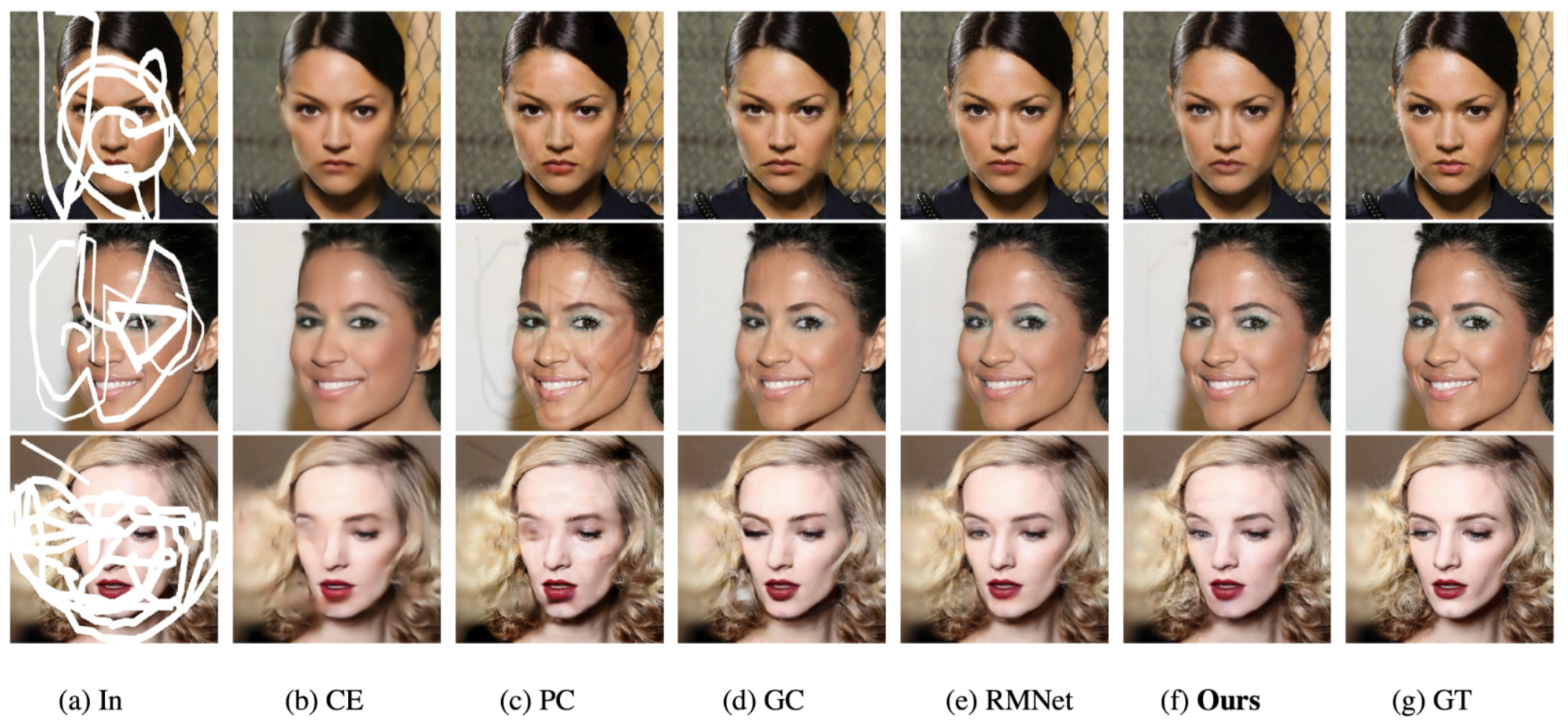} 
	\caption{Visual comparison of the inpainted results by our models \textbf{Ours},  \textbf{CE}, \textbf{PC}, \textbf{GC} and \textbf{RMNet} on CelebA-HQ \cite{liu2018image} where MaskDataset1 is used as masking method with mask hole-to-image ratios [0.01,0.6].}
	\label{fig:ynentresults}
\end{figure*}

The figures in this section depict the visual results of the V-LinkNet method. For comparison with the benchmarks and a variation of our model, we show the generated face images in Figure~\ref{fig:ynentresults}. It shows \textbf{CE} struggles with arbitrary hole-to-image mask regions and the generated image is blurry, while \textbf{PC} and \textbf{GC} leave a bit of artefacts (best viewed when zoomed) on the generated image. Focusing on the face and hair regions, our model performs better than the state of the art with no artefacts left on the inpainted regions. However, despite marginally comparable quantitative results on full inpainted images, our model completes and generates the facial image with no visible boundaries of the binary masks as seen on the generated images completed by the state of the art. 

%Figure~\ref{fig:ynentresults} depicts a visual comparison of the V-LinkNet method to the state of the art. 

\begin{figure}[!ht]   
	\centering
	\includegraphics[width=1.\linewidth]{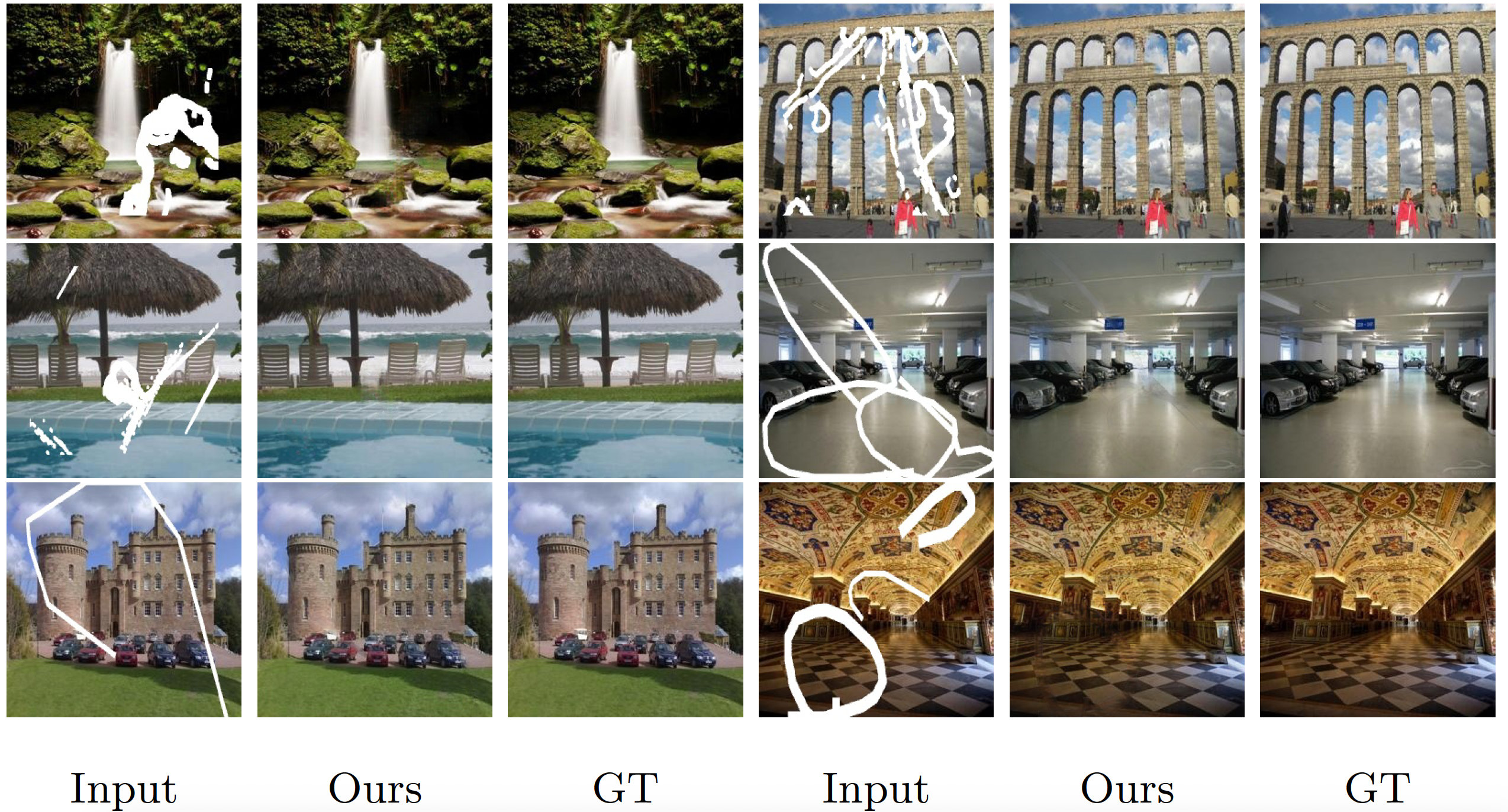} 
	\caption{Results showing inpainted images using V-LinkNet on Places2 Dataset with MaskDataset1 of our standardized test set ranging from [0.1-0.6], where images in column Input are the masked-image generated using the Quick-Draw Mask dataset \cite{iskakov2018semi}; images in column Ours are the results of inpainting by our proposed method; and images in column GT are the ground-truth.}
	\label{fig:Places2VLNetresults}
\end{figure}

\begin{figure} [!ht]  
	\centering
	\includegraphics[width=1.\linewidth]{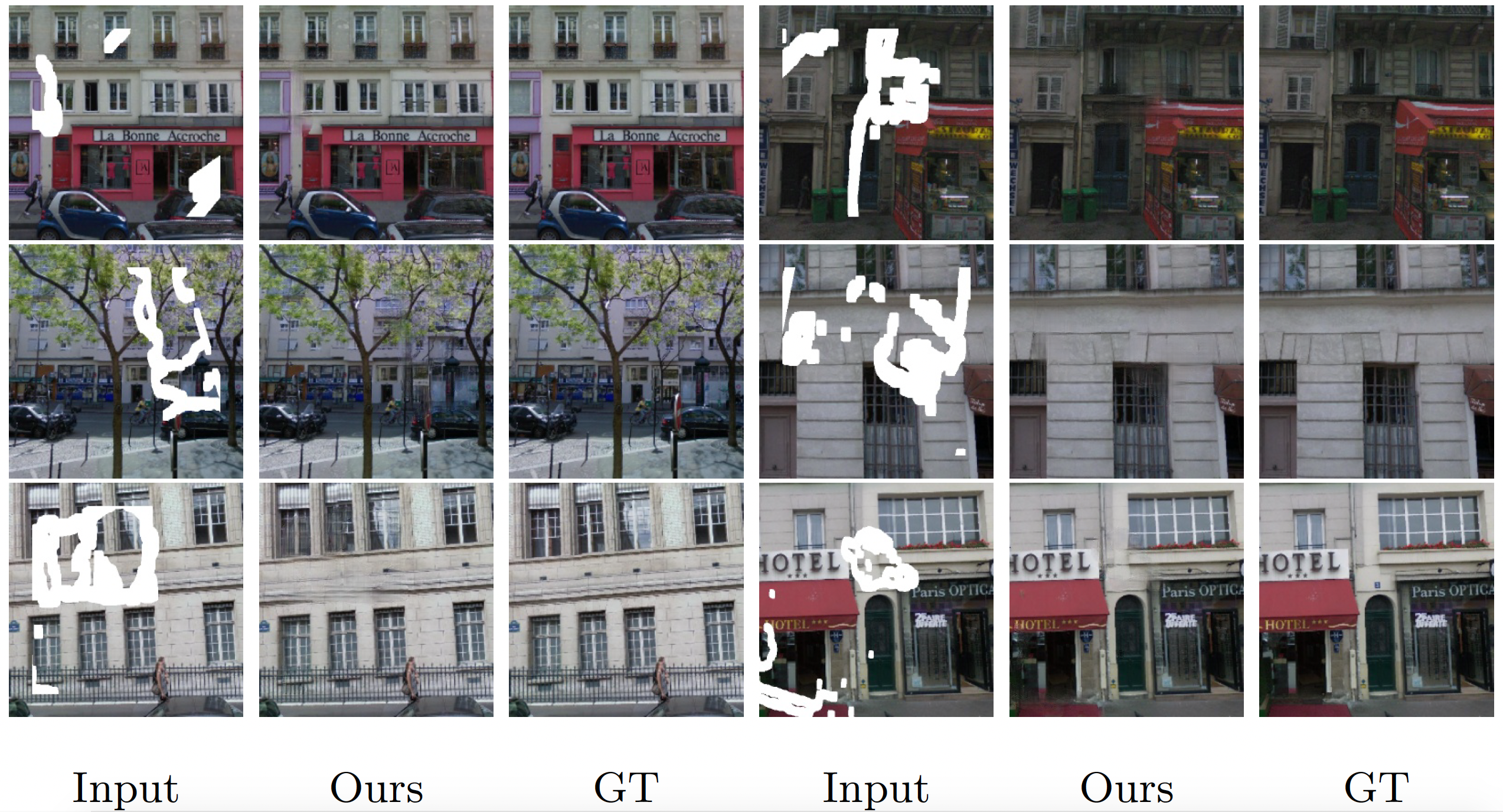} 
	\caption{Visual results showing inpainted images using V-LinkNet on Paris Street View Dataset with MaskDataset2 of our standardized test set, where images in column Input are the masked-image; images in column Ours are the results of inpainting by our proposed method; and images in column GT are the ground-truth.}
	\label{fig:PSVVLNetresults}
\end{figure}

The visual comparison of Places2 and Paris Street View datasets best represent how our model can generalise to natural scene images. The generated images are shown on Figure~\ref{fig:Places2VLNetresults} for Places2 \cite{zhou2017places} while Figure~\ref{fig:PSVVLNetresults} shows the inpainted images generated from Paris Street View dataset \cite{liu2020rethinking}.

\subsection{Quantitative Results}
It is important to note that the visual and semantic understanding of the completed regions is critical to the audience when inpainting in the wild. This is because the visual quality of the blending between the inpainted regions and the original unmasked regions should be unnoticeable in real-world scenarios.
However, in computer vision, we use quantitative evaluation to track model performance. Based on previous state-of-the-art research, we use the Mean Absolute Error (MAE), Frechet Inception Distance (FID), Peak Signal to Noise Ratio (PSNR), and SSIM to quantify performance against the state of the art (\cite{pathak2016context,liu2018image,yu2019free}). The high values obtained for MAE and FID show poor performance of the model whereas lower values for these metrics indicate better performance. For clarity, we have included in the table, where $\dagger$ indicates lower is better and $\uplus$ indicates higher is better. 
% PSNR and SSIM with higher values will indicate the prediction is closer to the ground-truth image, which will have a maximum score value of 1. 

Table \ref{tab:vgantable1} shows the quantitative evaluation for the inpainted images on CelebA-HQ testing set, with the best results in bold. Our proposed method achieved the best FID of 2.76 and SSIM of 0.96 when compared to the baseline models.

\begin{table}
	\begin{center}
		\caption{Quantitative comparison of various performance assessment metrics on 3,000 set 1 test images from the standard protocol of CelebA dataset. $\dagger$ indicates lower is better. $\uplus$ indicates higher is better.}
		\label{tab:vgantable1}
		\large
		\scalebox{0.6}{
			\begin{tabular}{|p{24mm}|p{35mm}|p{15mm}p{15mm}p{15mm}p{15mm}|}
				\hline
				\multicolumn{2}{|c|}{Performance Assessment}& &&&\\ \cline{1-2}
				%\cmidrule(r){1-2}
				Method & Author & MAE $\dagger$&FID$\dagger$&PSNR $\uplus$&SSIM $\uplus$\\
				\hline
				\textbf{CE}& Pathak et al. \cite{pathak2016context}&129.96&29.96&32.61&0.69\\
				\textbf{PC}&Liu et al. \cite{liu2018image}&98.01 &15.86 &33.03&0.81\\
				\textbf{GC}& Yu et al. \cite{yu2019free}&43.10 & 4.29 &39.96&0.92 \\
				\textbf{RMNet}&  Jam et al. \cite{jam2020r}&\textbf{31.91} & 3.09 &\textbf{40.40}&0.94 \\		
				% 		\textbf{VN1}&(V-LinkNet)& -- & 37.81&3.91 & 35.54 &0.92\\
				\textbf{V-LinkNet} &V-LinkNet &37.97& \textbf{2.76} & 39.75 &\textbf{0.96}\\
				\hline
			\end{tabular}
		}
	\end{center}
\end{table}

In addition, we perform quantitative measures on Places2 and Paris Street View datasets to test the ability of V-LinkNet in other image types. We compare the results to the state of the art \cite{jam2020r} and present the findings in Table~\ref{table:VGANresult3}. On the Paris Street View dataset, our proposed model outperformed the state of the art \cite{jam2020r} with SSIM of 0.95, but achieved marginally comparable result on the Places2 dataset, with SSIM of 0.91. The best results in Table~\ref{table:VGANresult3} are highlighted in bold.

\begin{table}[!htb]
	\begin{center}
		\caption{
			The inpainting results of V-LinkNet on Paris Street View and Places2, where our standard protocol MaskDataset1 \cite{iskakov2018semi} is used as masking method with mask hole-to-image ratios range between [0.01,0.6]. The results are compared against the state-of-the-art \cite{jam2020r}. $\dagger$ Lower is better. $\uplus$ Higher is better.
		}
		\label{table:VGANresult3}
		\large
		\scalebox{0.6}{
			\begin{tabular}{|p{35mm}|p{24mm}|p{15mm}p{15mm}p{15mm}p{15mm}|}
				\hline
				\multicolumn{2}{|c|}{Performance Assessment}& &&&\\ \cline{1-2}
				%\cmidrule(r){1-2}
				Dataset & Method & MAE $\dagger$&FID$\dagger$&PSNR $\uplus$&SSIM $\uplus$\\
				\hline
				\textbf{Paris Street View }& RMNet \cite{jam2020r})&33.81 & 17.64  & 39.55 & 0.91\\
				\textbf{Paris Street View}&V-LinkNet& \textbf{26.60} &\textbf{14.94}  & \textbf{40.9} & \textbf{0.95}\\\hline
				\textbf{Places2}& RMNet \cite{jam2020r} & \textbf{27.77}& \textbf{4.47}  & \textbf{39.66} & \textbf{0.93}\\
				\textbf{Places2}&  V-LinkNet &107.68  & 38.34 & 34.45 &0.91 \\		
				\hline
			\end{tabular}
		}
	\end{center}
\end{table}

\section{Ablation Study}
\label{sec:vnetablation}
\begin{figure}[!ht]
	\centering
	\includegraphics[width=1.\linewidth]{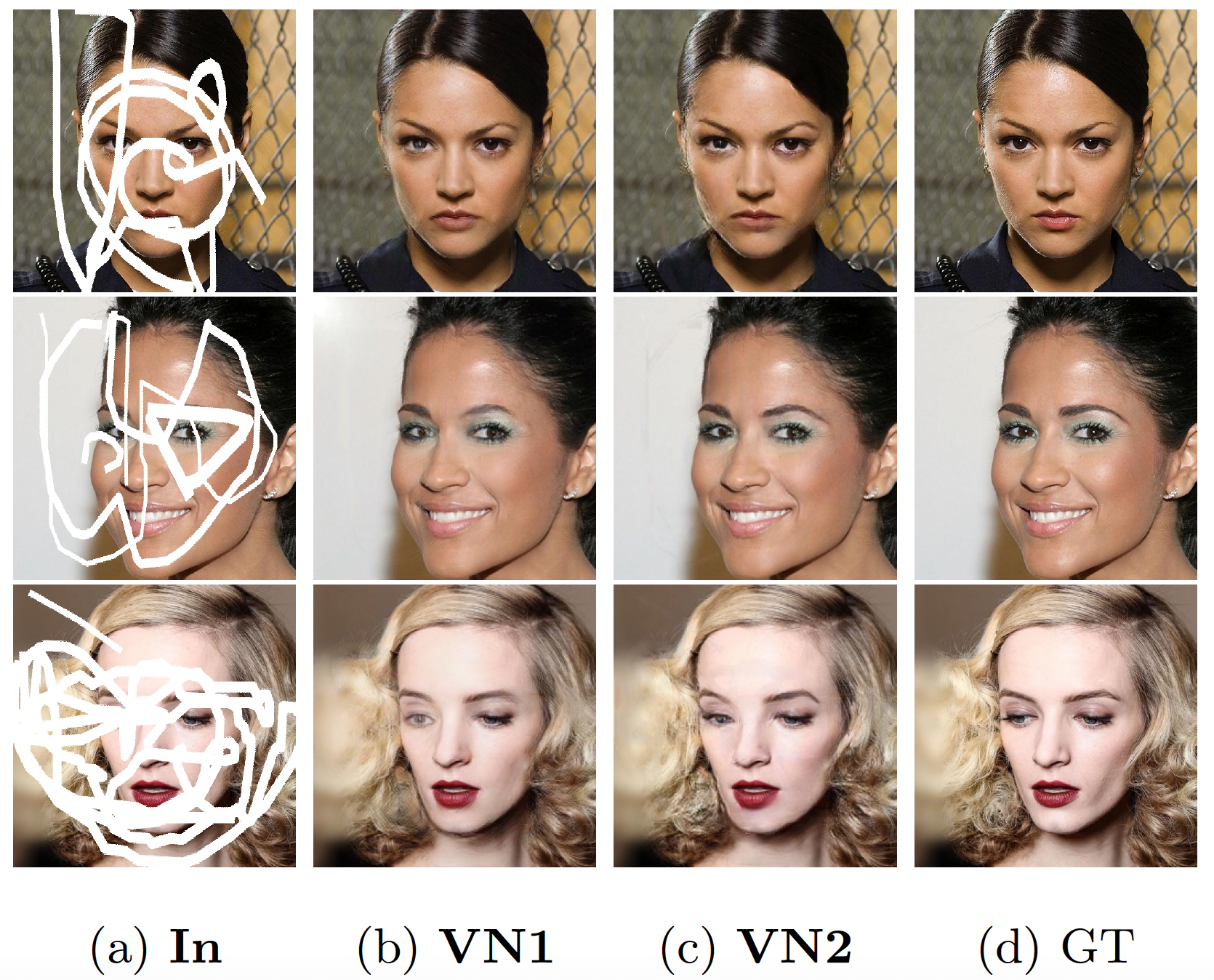} 
	\caption{For ablation study, we compare the inpainted results by variations our models \textbf{VN1}, \textbf{VN2}, on CelebA-HQ \cite{liu2018image} where MaskDataset1 from our standardised test set is used as masking method with mask hole-to-image ratios [0.01,0.6].}
	\label{fig:vnentablationresults}
\end{figure}

To understand the proposed method, an investigation is carried out to demonstrate the effectiveness of each component contributing to the image inpainting task. We carry out the ablation study V-LinkNet's performance on the proposed standard protocol of CelebA-HQ testing set. First, we evaluate the model using the latent space feature loss combined with the edge-based (Sobel) gradient loss. For the purpose of space on the Figure~\ref{fig:vnentablationresults}, we name this model as VN1, and then the full model with features losses and RSTL denoted by VN2. The visual results are shown on Figure~\ref{fig:vnentablationresults} and the quantitative evaluations on Table~\ref{tab:ablationvnettable1}. 

\subsection{Latent space feature loss combined with edge-based gradient loss (VN1)}
We slight modify the RSTL by removing the pooling unit. The modified layer is a residual block with the concept of attention in our inpainting task. We perform $1 \times 1$ convolutions on $g_{\theta_{A}}(\phi)$ and $g_{\theta_{B}}(\phi)$ output and concatenate the projected features maps. For dynamic feature selection, a softmax function is utilised on the concatenated feature map. Applying softmax after $1 \times 1$ convolutions on each encoder output enables precise feature values, thus preserving local and detailed information.

During the experiment, we use $L_{vgg}$, $L_{edgeLoss}$ combined with $L_{1}$ pixel-wise reconstruction loss.
We notice that using the $L_{\phi}$ loss combined with $L_{edgeLoss}$ gets rid of checker-board artefacts on the generated image. We notice that Sobel aids noise reduction and enhances the image quality of the generated output. 
% However, the quantitative evaluation of the model with this loss is not great compared to the full model without the Sobel loss.

\subsection{Full model with feature losses and RSTL (VN2)}
This section examines whether residual features from our recursive residual pooling unit has a positive effect on our model. The results in Table~\ref{tab:ablationvnettable1} demonstrate that residual refinement has a positive impact on the overall performance of our model. According to our findings, this improvement is attributable to the elimination of low-level information as a result of the pooling units being interconnected residually, which allows direct backpropagation of high-level information throughout the learning process.

\begin{table}
	\begin{center}
		\caption{Quantitative comparison of the performance assessment metrics on 3,000 images from the standard protocol test dataset of CelebA-HQ. The masking method applied is based on MaskDataset1. $\dagger$ Lower is better. $\uplus$ Higher is better.}
		\label{tab:ablationvnettable1}
		\large
		\scalebox{0.65}{
			\begin{tabular}{|p{24mm}|p{20mm}|p{15mm}p{15mm}p{15mm}p{15mm}|}
				\hline
				\multicolumn{2}{|c|}{Performance Assessment}& &&&\\ \cline{1-2}
				%\cmidrule(r){1-2}
				Method & Losses & MAE $\dagger$&FID$\dagger$&PSNR $\uplus$&SSIM $\uplus$\\
				\hline
				\textbf{(VN1)}& VN1 & \textbf{37.81}&3.91 & 35.54 &0.92\\
				\textbf{(VN2)} &VN2 &37.97 & \textbf{2.76} & \textbf{39.75} &\textbf{0.96}\\
				\hline
			\end{tabular}
		}
	\end{center}
\end{table}

\subsection{Quantitative evaluation of the standard protocol test set for facial images.}
This protocol is designed to evaluate the performance on a set of mask and images. The mask ratios in the Masksets range from [0.01,0.6]. The different MaskDataset and ratios are: MaskDataset1 [0.1,0.6], MaskDataset2 [0.01,0.1], MaskDataset3 [0.1,0.3], MaskDataset4 [0.3,0.4], MaskDataset5 [0.5,0.6] and MaskDataset6 [0.1,0.4]. 

\begin{table}[!ht]
	\begin{center}
		\caption{{Summary of quantitative results of the proposed standardised test set on CelebA-HQ \cite{karras2017progressive} and Paris Street View \cite{doersch2012makes} datasets using V-LinkNet. The performance evaluation vary from maskset to imageset and are approximated to 2 decimal places. The results included are for distortions (image-to-mask ratio) between 10\%-20\% on image sizes $256 \times 256$. $\dagger$ Lower is better. $\uplus$ Higher is better.}}
		\label{table:dataset}
		\large
		\scalebox{0.6}{
			\begin{tabular}{|p{45mm}|p{20mm}|p{12mm}p{12mm}p{12mm}p{12mm}|}
				\hline
				\multicolumn{2}{|c|}{Performance Assessment}& &&&\\ \cline{1-2}
				%\cmidrule(r){1-2}
				Dataset/Mask Ratio & Mask Type & MAE $\dagger$& FID$\dagger$&PSNR $\uplus$&SSIM $\uplus$\\
				\hline
				MaskDataset1 [0.01,0.6]& Irregular& 37.97 & \textbf{2.76}  & \textbf{39.75} & \textbf{0.96}\\
				MaskDataset2 [0.01,0.1] &Irregular & \textbf{21.35} & 3.36 &39.04 &0.94 \\
				MaskDataset3 [0.1,0.3] & Irregular&33.64& 5.23  &36.53 & 0.91\\	
				MaskDataset4 [0.3,0.4] &Irregular &64.15  &12.06 &33.72  & 0.89\\		
				MaskDataset5 [0.5,0.6] &Irregular & 107.33 &15.82 & 31.90 &0.74 \\		
				MaskDataset6 [0.1,0.4] &Irregular &25.75 &4.19 & 37.7 &0.93 \\
				\hline
			\end{tabular}
		}
	\end{center}
\end{table}

The MaskDataset6 are selected irregular masks that are used as masking method for more than one image (i.e one mask to many). Each mask is evaluated on more than one image and the performance is different across the dataset. The overall results are shown in Table~\ref{table:dataset}. The V-LinkNet has demonstrated overall best performance when presented with mask of various size ranges. 
% However, with tiny mask it demonstrates great performance and when presented with bigger masks, the results are not at its best.

This study is conducted to identify biases for different masks on different images and propose a standard protocol that will propel research in image inpainting. The mask-to-area ratio was determined using OpenCV toolbox. Based on this study, we observed that the performance of an algorithm will very much depend on the mask type and the image type. There are some conditions on a facial image that can influence the performance such as pose, lighting, features and background. In the case of CelebA-HQ dataset, we observed that if the mask is on the skin region, the performance evaluation has better scores compared to when the mask is on the a difficult background with variations in lighting conditions. Furthermore, the mask applied to a face posed at an angle will influence the results as shown in Figure~\ref{fig:v-gan_eval_faces} second row, second set of images. Based on this finding, we proposed standardised test sets to support a fairer comparison in future research.

\section{Conclusion}

% After each convolution, we use ELU as the activation function, followed by maxpooling and batch normalization. To avoid exploding gradients, l2 regularizers at 1E-5 are used within the convolution blocks. 

We proposed V-LinkNet, a novel image inpainting technique that uses two encoders to learn from each other, which advanced the field by outperformed previous methods. To tackle the irregular-holes inpainting problem, we presented a dual-encoder method that exploits semantic coherency across textural features through enforced collaboration. For each spatial location, each convolutional layer expects a certainty. V-LinkNet handles high-level feature propagation as a learned operation within a residual unit designed with maxpooling units and a residual convolution unit to create the full layer. The proposed solution is simple and efficient, and it acts as a bridge to the decoder. 

We presented a RSTL that serves as a propagation module to optimise the projected textures from features of both encoders in a morphological manner. This module provides consistency and coherency by combining two features into one. The V-LinkNet model can propagate high-level features to the decoder using this unit. To validate its efficacy, we conducted an ablation study with various model components. The unmodified RSTL combined with the feature losses loss is found to be the best model combination. We contend that the combined recursive residual unit, which is linked to residual pooling and residual convolution, enables direct backpropagation within the bottleneck's deeper layers. This forces the selection of high-level information during decoder layer propagation, resulting in high quality reconstruction of inpainted regions.  %Certainty, 

From the results, we observed that the model achieves learning of high-level features with propagation to decoding layers. Furthermore, the feature-wise loss model shared by both encoders aids the model during early learning, resulting in a better learning strategy shared by both encoders. The losses and Wasserstein discriminators improve the semantic consistency of our model, which ensures fine contextual information. Our approach successfully generate quality semantic structural and textural features that match the ground-truth image. Our research provides new insights on the need of a standard protocol, where we shared this protocol as a recommendation for performance evaluation.

% \section*{Acknowledgments}
% This should be a simple paragraph before the References to thank those individuals and institutions who have supported your work on this article.

% {\appendix[Proof of the Zonklar Equations]
% Use $\backslash${\tt{appendix}} if you have a single appendix:
% Do not use $\backslash${\tt{section}} anymore after $\backslash${\tt{appendix}}, only $\backslash${\tt{section*}}.
% If you have multiple appendixes use $\backslash${\tt{appendices}} then use $\backslash${\tt{section}} to start each appendix.
% You must declare a $\backslash${\tt{section}} before using any $\backslash${\tt{subsection}} or using $\backslash${\tt{label}} ($\backslash${\tt{appendices}} by itself
%  starts a section numbered zero.)}


% \section{References Section}
% You can use a bibliography generated by BibTeX as a .bbl file.
%  BibTeX documentation can be easily obtained at:
%  http://mirror.ctan.org/biblio/bibtex/contrib/doc/
%  The IEEEtran BibTeX style support page is:
%  http://www.michaelshell.org/tex/ieeetran/bibtex/

%  % argument is your BibTeX string definitions and bibliography database(s)
% %\bibliography{IEEEabrv,../bib/paper}
%

\section*{Acknowledgment}
This work was supported by The Royal Society UK (INF \textbackslash PHD \textbackslash 180007 and IF160006). We gratefully acknowledge the support of NVIDIA Corporation with the donation of the Quadro P6000 used for this research.

\bibliographystyle{IEEEtran}
\bibliography{refs.bib}

\end{document}